# Fast algorithm for Multiple-Circle detection on images using Learning Automata

Erik Cuevas[a], Fernando Wario[a], Valentín Osuna-Enciso[b], Daniel Zaldivar[a] and Marco Pérez-Cisneros[a]

[a]Departamento de Ciencias Computacionales
Universidad de Guadalajara, CUCEI
Av. Revolución 1500, Guadalajara, Jal, México
{[1]erik.cuevas, fernando.wario, daniel.zaldivar, marco.perez}@cucei.udg.mx

[b]Centro de Investigación en Computación-IPN
Av. Juan de Dios Batiz S/N
Col. Nueva Industrial Vallejo, Mexico, D. F. MEXICO
osuna@cic.ipn.mx

**Abstract**

Hough transform (HT) has been the most common method for circle detection exhibiting robustness but adversely demanding a considerable computational load and large storage. Alternative approaches include heuristic methods that employ iterative optimization procedures for detecting multiple circles under the inconvenience that only one circle can be marked at each optimization cycle demanding a longer execution time. On the other hand, Learning Automata (LA) is a heuristic method to solve complex multi-modal optimization problems. Although LA converges to just one global minimum, the final probability distribution holds valuable information regarding other local minima which have emerged during the optimization process. The detection process is considered as a multi-modal optimization problem, allowing the detection of multiple circular shapes through only one optimization procedure. The algorithm uses a combination of three edge points as parameters to determine circles candidates. A reinforcement signal determines if such circle candidates are actually present at the image. Guided by the values of such reinforcement signal, the set of encoded candidate circles are evolved using the LA so that they can fit into actual circular shapes over the edge-only map of the image. The overall approach is a fast multiple-circle detector despite facing complicated conditions.

1. Introduction

The problem of detecting circular features holds paramount importance for image analysis applications such as automatic inspection of manufactured products and components, aided vectorization of drawings, target detection, etc. [1]. The circle detection in digital images is commonly done by applying the Circular Hough Transform [9] (CHT). A typical Hough-based approach employs an edge detector and uses edge information to infer locations and radius values. Peak detection is then performed by averaging, filtering and histogramming the transform space. However, such approach requires a large storage space given the 3-D cells needed to cover the parameters ($x$, $y$, $r$), the computational complexity and the low processing speed. Moreover, the accuracy on the detected-circle parameters is poor, particularly under noisy conditions [10]. For digital images of significant width and height plus densely populated edge pixels, the required processing time for CHT makes it prohibitive to be deployed in real time applications. In order to overcome such a problem, researchers have proposed new approaches considering methods which clearly differ from CHT. Some examples are the Probabilistic Hough transform [11], the randomized Hough transform (RHT) [12] and the fuzzy Hough transform [13]. On the other hand, in [14], Lu & Tan have recently proposed a novel approach based on RHT, also known as the Iterative Randomized Hough Transformation (IRHT) which has delivered evidence of students achieving better results on complex images and noisy environments. The algorithm iteratively applies the randomized Hough transform (RHT) to a region of interest within the image which is determined from the latest estimation of ellipse/circle parameters.

Circle detection in computer vision has also been approached as an optimization problem as an alternative to Hough Transform-based techniques. Ayala–Ramirez *et al* presented a GA based circle detector in [15] which is capable of detecting circles on real images, but frequently missing imperfect circles. On the other hand, Dasgupta et al. [16] have recently proposed an automatic circle detector (BFAOA) using the Bacterial Foraging Algorithm as an optimization method. However, both methods demand the algorithm's repeated execution in order to detect several circular shapes, i.e. one procedure per circle. Therefore, only one circle can be detected at each optimization cycle yielding a longer execution time for images holding several circular shapes. A different approach is presented in [17] where the Clonal

 

Selection Algorithm (CSA) is modified in order to detect multiple-circles, assuming that the overall detection process is a multi-modal optimization problem.

The main motivation behind the use of LA [18] as optimization algorithm for parameter estimation emerges naturally from its capabilities to operate with multimodal functions. Although the automaton would eventually converge to only one global minimum, the algorithm holds important information regarding local minima as the optimization process evolves.

The LA method searches within the probability space rather than exploring the parameter space as commonly done by other optimization techniques [19]. LA method does not need knowledge of the environment or any other analytical reference to the function to be optimized. In order to select future search points, LA builds and updates a probability density function using the reinforcement signal (objective function) and the learning algorithm.

Over recent years, LA has been successfully applied to solve different sorts of engineering problems, including pattern recognition [20], signal processing [21], adaptive control [22], power systems [23] and image processing [24] among others. Recently, several effective LA-based algorithms have been proposed for multimodal complex function optimization [21, 25, 26, 27] yielding either an equivalent or better experimental performance in comparison to similar GA methods [26].

This paper presents an algorithm for the automatic detection of multiple circular shapes embedded into complicated noisy images with no consideration of conventional Hough transform principles. The overall detection process is accomplished by an LA-based optimization approach that searches circular shapes over the entire edge-map. A combination of three non-collinear edge points evaluates some candidate circles (actions) within the edge-only image of the scene, while a reinforcement signal (matching function) is used to measure the existence of a candidate circle. Guided by the values of such reinforcement signal, the set of encoded candidate circles are evolved using the LA so that the best candidate can fit into an actual circle within the edge-only image. Just after the optimization step is finished, the probabilistic distribution is studied for locating and marking significant local minima (remaining circles). Experimental evidence shows the effectiveness of the method for detecting multiple circles under complicated conditions. A comparison with one state-of-the-art GA-based method [15], the BFAOA [16], the CSA detector [17], and the RHT technique [12,14] on different images has been included to demonstrate the performance of the proposed approach. Conclusions of the experimental comparison are validated through statistical tests that support the discussion suitably.

The paper is organized as follows: Section 2 introduces the LA method while Section 3 presents the overall LA-based circle detector. The multiple circle detection procedure is then presented by Section 4 with experimental results for the proposed approach being discussed in Section 5. Section 6 offers some comments and conclusions regarding the algorithm's performance.

**2. Learning Automata.**

LA operates by selecting actions stochastically within an environment and assessing a measure of the system performance. Figure 1a shows the typical learning system architecture. The automaton selects probabilistically an action $x_i$ from the set of actions. After applying such action to the environment, a reinforcement signal $\beta(x_i)$ is provided through the evaluation function. The internal probability distribution is updated whereby actions that achieve desirable performance are reinforced via an increased probability while those underperforming actions are penalized or left unchanged depending on the particular learning rule which has been employed. The average performance of the system must improve over the time until a performance index is reached. In terms of optimization problems, the action with the highest probability would correspond to the minimum as demonstrated by rigorous proofs of convergence available in [19] and [28].

A wide variety of learning rules have been reported in the literature. However, one of the most widely used algorithms is the linear reward/inaction ($L_{RI}$) scheme which has shown effective convergence properties (see [29]). Considering an automaton $A$ with $n$ different actions, where $x_i$ represents the action $i$ of a set of $n$ possible actions, the response to an action $x_i$, at time step $k$, yields a probability updating rule as follows:

$$p_i(k+1) = p_i(k) + \theta \cdot \beta(x_i) \cdot (1 - p_i(k)) \tag{1}$$

$$p_j(k+1) = p_j(k) - \theta \cdot \beta(x_i) \cdot p_j(k), \text{ if } j \neq i$$

with $\theta$ being the learning rate and $0 < \theta < 1$, $\beta(\cdot) \in [0,1]$ being the reinforcement signal whose value $\beta(\cdot) = 1$ indicates the maximum reward and $\beta(\cdot) = 0$ signals a null reward considering $i, j \in \{1, \ldots, n\}$. Eventually, the probability of successful actions will increase to become close to unity. In case a single and foremost successful action prevails, the automaton is deemed to have converged.

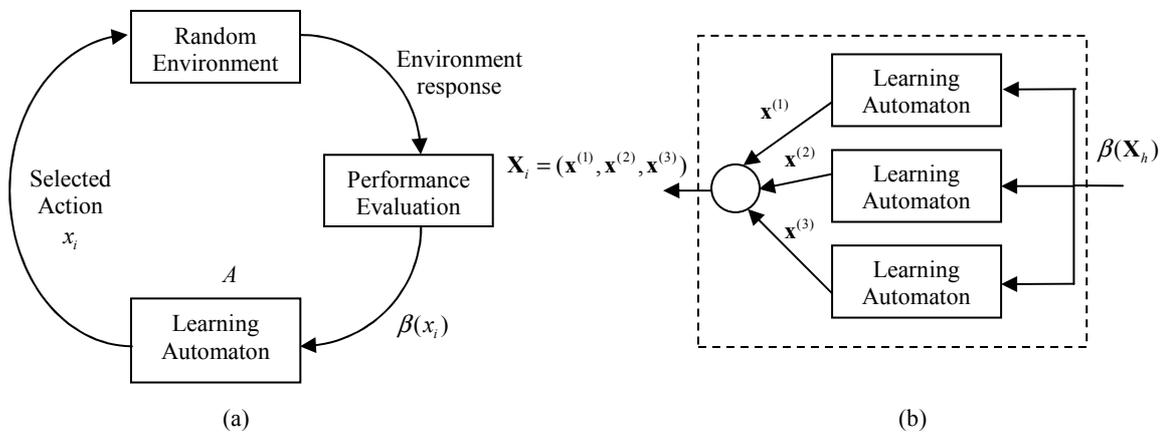

**Fig. 1.** (a) Reinforcement learning system and (b) the interconnected automata

In order to solve multidimensional problems, the learning automata can also become connected within a parallel setup (see Figure 1b). Each automaton operates with a simple parameter and while the concatenation allows working within the multidimensional space. There is no inter-automata communication as the only joining path is through the environment. In [21], it is shown how discrete stochastic learning automata can be used to determine the global optimum for problems with multi-modal surfaces.

## 3. Circle detection using LA.

*3.1. Data preprocessing.*

At this work, a circular shape is represented by a well-known second degree equation (Equation 10) whose parameters represent three non-collinear points matching a given circular shape. In order to apply the function, the candidate images must be preprocessed by the well-known Canny edge detection algorithm yielding a edge-only image. The segmentation considers a single-pixel contour and stores locations for each edge point in a vector array $P_t = \{p_1, p_2, \ldots, p_{N_t}\}$ with $N_t$ being the total number of edge pixels.

The algorithm stores the $(x_i, y_i)$ coordinates for each edge pixel $p_i$ in the edge vector $P_t$. Following the RHT technique in [12], only a representative percentage of edge points (around 5%) are considered for building a new vector array $P = \{p_1, p_2, \ldots, p_{N_p}\}$, where $N_p$ is the number of edge pixels randomly selected from $P_t$. Such points are the only potential candidates to define circles by considering triplets.

In order to construct each circle candidate (or actions within the LA framework), the indexes $i_1$, $i_2$ and $i_3$ which represents the three edge points, must be combined assuming that the circle's contour connects the points $p_{i_1}$, $p_{i_2}$ and $p_{i_3}$. Therefore, the circle passing over such points may be considered as a potential solution to the detection problem. The number of actions, which is represented by $n_c$, gathers prototype circles whose three points are grouped into vector $P$. All the LA actions are circles holding a radius within the interest limits. Very small circles are cancelled as they may represent noise.

The algorithm ensures that each circle under consideration yields a different action. For instance, two or more actions corresponding to the same circle must avoid competing among themselves reducing the number of decisions and the required computational time.

The LA solution is based on tracking the probability evolution of each circle candidate -also known as actions, as they are modified according to their actual affinity. This affinity is computed using an objective function which determines if a circle candidate is actually present in the image. A given circle candidate showing the highest probability measure after a number of cycles may be assumed as a circle actually present in the image.

Although the Hough Transform methods for circle detection also use three edge points to cast one vote for a potential circular shape in the parameter space, they seriously require huge amounts of memory and longer computational times to reach a sub-pixel resolution. On the contrary, the LA method employs an objective function to yield improvement at each generation step, discriminating among non-plausible circles and avoiding unnecessary testing of certain image points. However, both methods require a compulsory evidence-collecting step for future iterations.

*3.2. Action representation.*

Each action (circle) $C_l$ of the automaton is defined by the combination of three edge points (where $l$ represents the action number or circle candidate). Under such representation, edge points are stored by assigning an index which refers to their relative position within the edge array $P$. It is necessary to encode the action as a circle which passes through the three edge points $p_i$, $p_j$ and $p_k$ ($C_l = \{p_i, p_j, p_k\}$), so each circle $C_l$ will be represented by three parameters $x_0$, $y_0$ and $r$, considering $(x_0, y_0)$ as the coordinates of the center of the circle and $r$ its radius. The circle passing through the three edge points can thus be modeled as follows:

$$(x - x_0)^2 + (y - y_0)^2 = r^2 \tag{2}$$

where $x_0$ and $y_0$ are calculated using the following equations:

$$x_0 = \frac{\det(\mathbf{A})}{4((x_j - x_i)(y_k - y_i) - (x_k - x_i)(y_j - y_i))}, \quad y_0 = \frac{\det(\mathbf{B})}{4((x_j - x_i)(y_k - y_i) - (x_k - x_i)(y_j - y_i))}, \tag{3}$$

with $\det(\mathbf{A})$ and $\det(\mathbf{B})$ representing the determinants of matrices $\mathbf{A}$ and $\mathbf{B}$ respectively, whose values are defined as follows:

$$\mathbf{A} = \begin{bmatrix} x_j^2 + y_j^2 - (x_i^2 + y_i^2) & 2 \cdot (y_j - y_i) \\ x_k^2 + y_k^2 - (x_i^2 + y_i^2) & 2 \cdot (y_k - y_i) \end{bmatrix} \quad \mathbf{B} = \begin{bmatrix} 2 \cdot (x_j - x_i) & x_j^2 + y_j^2 - (x_i^2 + y_i^2) \\ 2 \cdot (x_k - x_i) & x_k^2 + y_k^2 - (x_i^2 + y_i^2) \end{bmatrix}, \tag{4}$$

the radius $r$ can therefore be calculated using:

$$r = \sqrt{(x_0 - x_d)^2 + (y_0 - y_d)^2}, \tag{5}$$

where $d \in \{i, j, k\}$, and $(x_d, y_d)$ are the coordinates of any of the three selected points defining the action. Figure 2 illustrates the parameters defined by Equations (3)-(5).

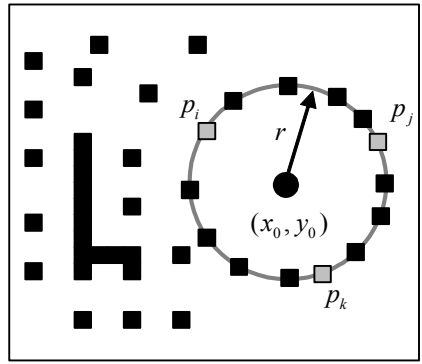

**Fig. 2.** Circle candidate (action) as it is shaped by the combination of the points $p_i$, $p_j$ and $p_k$.

The shaping parameters for the circle, $[x_0, y_0, r]$ can be represented as a transformation $T$ of the edge-vector's indexes $i, j$ and $k$.

$$[x_0, y_0, r] = T(i, j, k) \tag{6}$$

with $T$ being the transformation matrix from previous computations of $x_0$, $y_0$, and $r$.

*3.3 Reinforcement signal $\beta$.*

Optimization refers to the search of parameters that minimize or maximize an objective function or error expression. In order to calculate the error produced by a candidate solution $C_l$, the circumference

 

coordinates are calculated as a virtual shape which, in turn, must also be validated, i.e. if it really exists in the edge image. The test set is represented by $S_l = \{s_1, s_2, \ldots, s_{N_s}\}$, where $N_s$ are the number of points over which the existence of an edge point, corresponding to $C_l$, should be tested.

The test set $S$ is generated by the midpoint circle algorithm (MCA) [30]. It is a well-known algorithm to determine the required points for drawing a circle which requires only the radius $r$ and the center point $(x_0, y_0)$ as input arguments. The algorithm considers the circle equation $x^2 + y^2 = r^2$, with only the first octant. It draws a curve starting at point $(r, 0)$ and proceeds upwards-left by using integer additions and subtractions. The MCA aims to calculate the required points $S$ in order to represent the circle candidate. Despite this algorithm is considered as the quickest providing a sub-pixel precision, it is important to assure it does not consider points lying outside the image plane in $S$.

The reinforcement signal $\beta(C_l)$ represents the matching error produced between the pixels $S$ of the circle candidate $C_l$ (action) and the pixels that actually exist in the edge image, yielding:

$$\beta(C_l) = \frac{\sum_{i=1}^{N_s} E_l(x_i, y_i)}{N_s} \quad (7)$$

where $E_l(x_i, y_i)$ is a function that verifies the pixel existence in $(x_i, y_i)$, being $(x_i, y_i) \in S_l$ and $N_s$ is the number of pixels lying on the perimeter corresponding to $C_l$, currently under testing. Hence, function $E_l(x_i, y_i)$ may be defined as:

$$E_l(x_i, y_i) = \begin{cases} 1 & \text{if the pixel } (x_i, y_i) \text{ is an edge point} \\ 0 & \text{otherwise} \end{cases} \quad (8)$$

A value near to one of $\beta(C_l)$ implies a better response from the "circularity" operator. The LA algorithm is configured for a pre-selected cycle limit that is usually set to half of the number of actions previously gather within the automaton.

There are two cases for defining a solution (also known as action): first, the case of one action (circle candidate) generating a matching error under a pre-established limit. Second, the highest probability action is taken at the end of the learning process.

*3.4. LA implementation.*

The procedure of LA methodology can be summarized in the following steps:

**Step 1:** Apply the Canny filter to the original image.

**Step 2:** Select 5% of the edge pixels to build the *P* vector.

**Step 3:** Generate all feasible actions ($n_{all}$) through the combinations of *P*.

**Step 4:** Calculate $[x_0, y_0, r] = T(i, j, k)$ for all possible actions $n_{all}$. Then, eliminate actions that correspond to very small or repeated circles. The remaining actions ($n_c$) are used to set the final automaton. Finally, *kmax* must be set to $n_c / 2$.

**Step 5:** Set iteration *k*=0.

**Step 6:** Initially, set the same probability value for the $n_c$ actions, i.e $p(A_i, k) = 1/n_c$, with $A_i \in (A_1, A_2, \ldots A_{n_c})$.

**Step 7:** Repeat while $k < (n_c / 2)$

**Step 8:** Using a pseudo-random generator, obtain a *z* number between 1 and 0.

**Step 9:** Select $A_v \in (A_1, A_2, \ldots A_{n_c})$, considering that the area under the probability density function is $\sum_{i=1}^{v} p(A_i, k) > z$.

**Step 10:** Evaluate the affinity calculating $\beta(A_v)$ [Eq. (7)].

**Step 11:** Update the automaton's internal probability distribution using Eq.(1).

**Step 12:** Increase *k*, and jump to step 7.

**Step 13:** end of while

**Step 14:** The solution is $A_m$ representing the element (circle) with the highest probability, drawing the circle over the image.

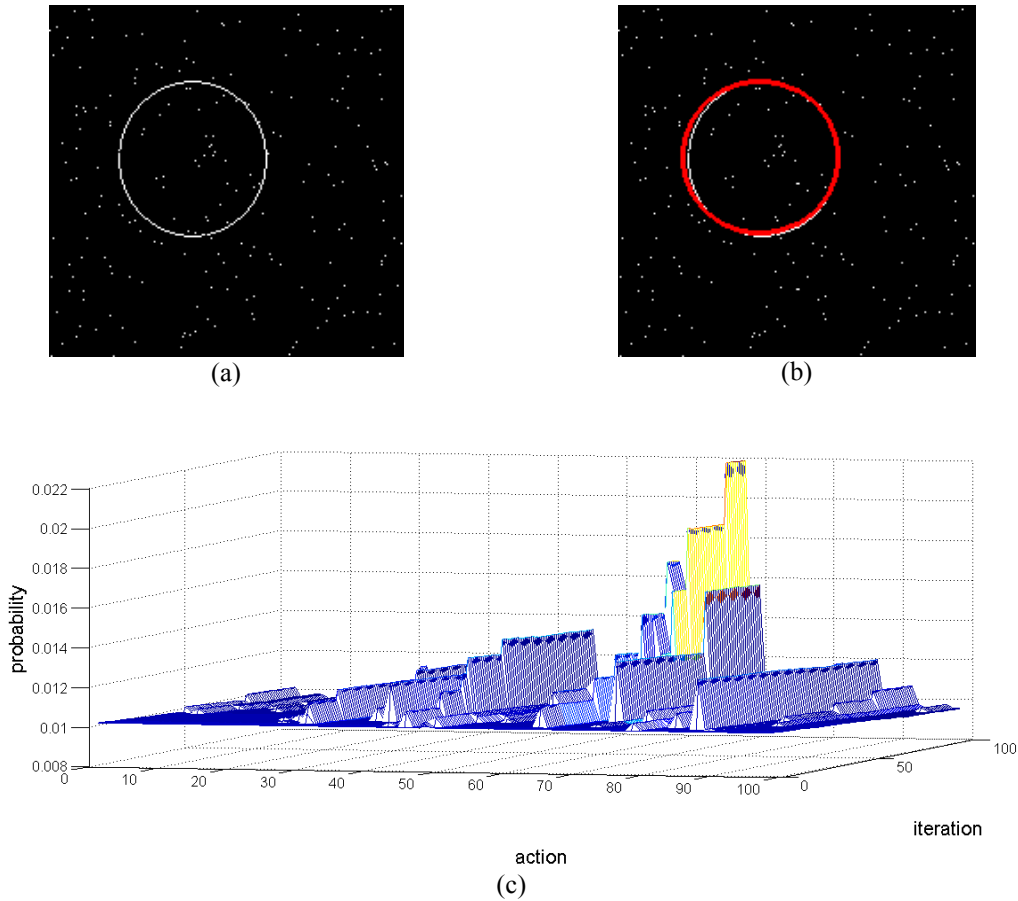

**Fig. 3.** Circle detection and the evolution of the probability density parameters. (a) Original image. (b) The detected circle is shown as an overlay, (c) Probability density chart after the algorithm's evolution.

Figure 3 shows one of the results produced by the LA-based detector. The input is the image at Figure 3a which features a noisy image of 200 x 200 pixels holding only one imperfect circle (ellipsoidal shape). The best matching circle is shown by Figure 3b after 50 cycles. For this experiment, the detection time has been 0.09 seconds. Figure 3c presents the action's probability distribution evolution with the highest peak representing the highest probability action.

**4. The multiple circle detection procedure.**

Other methods have been published in the literature for detecting multiple circular shapes in an image [15, 16]. However, such algorithms commonly find only one circular shape at time forcing a multiple application of the procedure in order to detect more than one circle. Nevertheless, the method proposed in this paper is able to detect single or multiples circles through only one optimization procedure. Guided by the values of a matching function (reinforcement signal), the set of encoded candidate circles are evolved using the LA so that the best circle candidate can fit into an actual circle within the edge-only image. Here, the best circle candidate means the action which has the high value according to the probability distribution. In order to detect other remaining circles in the image, the algorithm analyses the resultant probability distribution seeking other significant local minima.

In order to find other local minima, the probability distribution is arranged into a descending order. Then, the algorithm explores each action, one at a time, validating if such feature represents an actual circle. As several actions can represent the same circle, a distinctiveness factor $E_{s_{di}}$ is required to measure the mismatch between two given circles (actions). This distinctiveness factor is defined as follows:

$$E_{s_{di}} = |x_A - x_B| + |y_A - y_B| + |r_A - r_B| \qquad (9)$$

being $(x_A, y_A)$ and $r_A$ the central coordinates and radius of the circle $C_A$ respectively, while $(x_B, y_B)$, $r_B$ are the corresponding parameters of the circle $C_B$. A threshold value $E_{s_{TH}}$ is also calculated to decide whether two circles must be considered as different. $E_{s_{TH}}$ is thus computed as follows:

$$E_{s_{TH}} = \frac{r_{max} - r_{min}}{s} \qquad (10)$$

where $[r_{min}, r_{max}]$ is the feasible radii range and $s$ is a sensitivity parameter. By using a high $s$ value, two very similar circles would be considered as different while a smaller value for $s$ would consider them as similar circles.

Once the probability distribution has been obtained and arranged, the highest value $\Pr_{high}$ is assigned to the first circle. By exploring the remaining values, other circles are detected considering the discrimination rules sketched by Equations (9) and (10). The process is repeated until the action's probability reaches a minimum threshold $\Pr_{th}$. According to such threshold, the values above $\Pr_{th}$ represent significant actions (circles) and lower values are considered as false circles which are not contained within the image. After several experiments the value of $\Pr_{th}$ is set to $\Pr_{high}/10$.

The multiple circle detection procedure can be summarized in the following steps:

| | |
|---|---|
| **Step 1** | The parameter of sensitivity *s* is set in order to define $E_{s_{TH}}$. |
| **Step 2** | The actions (circle candidates) are organized in descending order according to their probabilities. |
| **Step 3** | The action with the highest probability ($\Pr_{high}$) is identified as the first circle $C_1$. |
| **Step 3** | The distinctiveness factor $E_{s_{di}}$ of circle $C_m$ (action *m*) with the next highest probability is evaluated with respect to $C_{m-1}$. If $E_{s_{di}} > E_{s_{TH}}$, then it is considered $C_m$ as a new circle otherwise the next action is evaluated. |
| **Step 4** | The step 3 is repeated until the action's probability reaches $\Pr_{high}/10$. |

## 5. Experimental results.

Experimental tests have been developed in order to evaluate the performance of the multiple-circle detector under different circumstances. The experiments address the following tasks:

(5.1) Multiple circle detection

(5.2) Circular approximation

(5.3) Overlapped circles

(5.4) Shape discrimination

(5.5) Incremental noise

(5.6) Performance comparison

Table 1 presents the parameters of the LA algorithm at this work. They have been kept in all test images after being experimentally defined.

| $kmax$ | $\theta$ | $r_{max}$ | $r_{min}$ | $s$ |
|---|---|---|---|---|
| $n_c/2$ | 0.001 | 150 | 40 | 2 |

**Table 1.** LA detector parameters applied to all tests after they have been experimental defined.

All the experiments have been implemented over a Pentium IV 2.5 GHz computer under C language programming. All the images are preprocessed by the standard Canny edge-detector using the image-processing toolbox for MATLAB R2008a.

*5.1. Multiple circle detection.*

The experimental setup includes two different images containing multiple circles of different size. Once the edge map is generated, some combinations of edge points are considered as the LA's set of actions. The Automata's learning process is thus executed until the number of cycles is reached (the maximum number of cycles is defined as 100) or one of the actions (first circle) probability is big enough to be considered as a global minimum (For all experiments a probability of 0.2 is considered). Other circles can also being detected following the analysis of the final probability distribution.

5.1.1. Natural images.

The first experimental batch tests the circle detection over natural images as those shown by Figure 4 including its final probability distribution (Figure 4a) and its correspondent descendent arrangement (Figure 4b) considering thirty most significant actions according to $\Pr_{high}/10$. The edge-map image is presented by Figure 4c while Figure 4d features the detected circles.

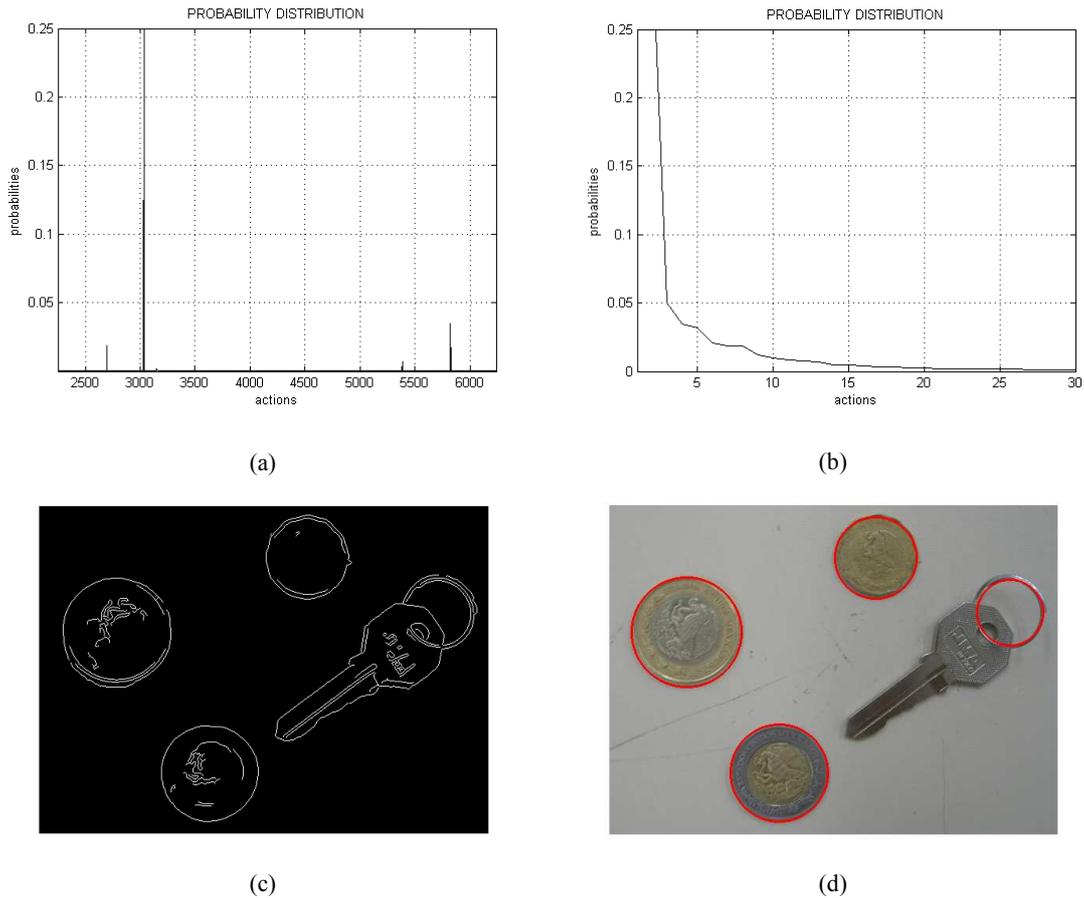

**Fig. 4.** An example of multiple circle detection. (a) The final probability distribution, (b) Thirty significant actions showing probability at descending order. (c) The edge-map image and (d) its correspondent detected circles.

5.1.2 Synthetic images.

The experimental setup includes the use of several synthetic images of different resolutions. Each image has been generated drawing multiple semi-circles, randomly located. Some images have been contaminated by adding noise to increase the complexity in the detection process. For all cases the algorithm is able to detect circles despite noise. The detection is robust to translation and scaling keeping a reasonably low elapsed time, typically under 1ms. Figure 5 shows the resulting analysis for one image taken out from the set.

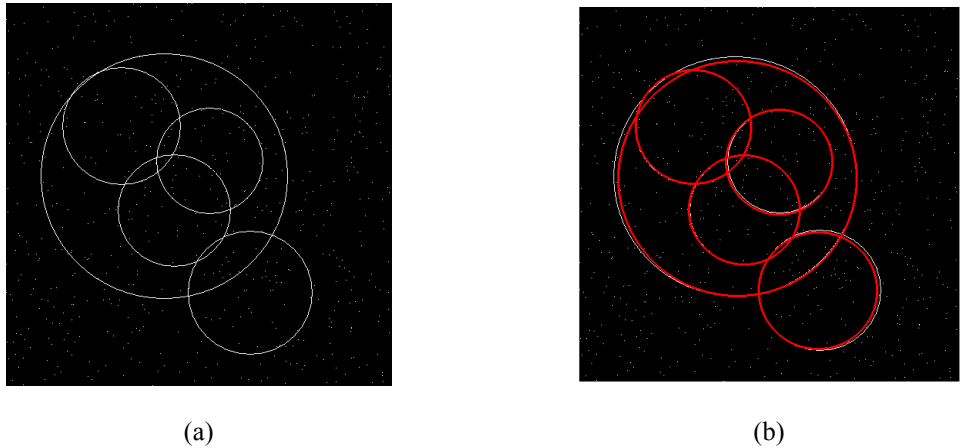

(a)  (b)

**Fig. 5.** Multiple circle detection over a synthetic image. (a) Synthetic image and (b) detected circles.

*5.2 Circular approximation.*

Since circle detection has been considered an optimization problem, it is possible to approximate a given shape as a circle concatenation and the LA method may be used to such purpose. The LA algorithm can continually find circles approximating a given shape according to their values on the final probability density. Figure 6 shows an approximation example after using the LA detector.

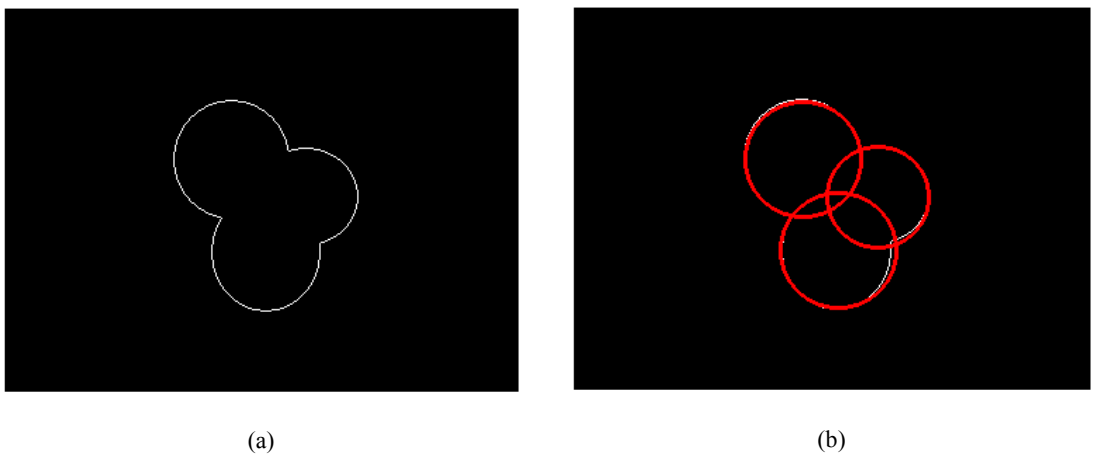

(a)  (b)

**Fig. 6.** Approximation of a partial circular shape by means of circle concatenation. (a) The edge-map and (b) the circle approximation

 

*5.3 Overlapped circles.*

Identifying overlapped circles within an image represents an important test to verify the performance of a circle detector. For the method reported in [15] and [16], the ability to detect overlapped circles is limited because of the amount of relevant information which is commonly lost during the masking process. On the contrary, the method at this paper does not eliminate pixels so they can be effectively considered in the detection stage. Our approach has been tested over a set of challenging images containing overlapped circles. Figure 7 shows the analysis for one of them, including its final probability distribution (Figure 7a) and its correspondent descendent arrangement (Figure 7b) which considers the 30 most significant actions according to $\Pr_{high}/10$). The edge-map image is presented by Figure 7c while Figure 7d shows the resulting detected circles.

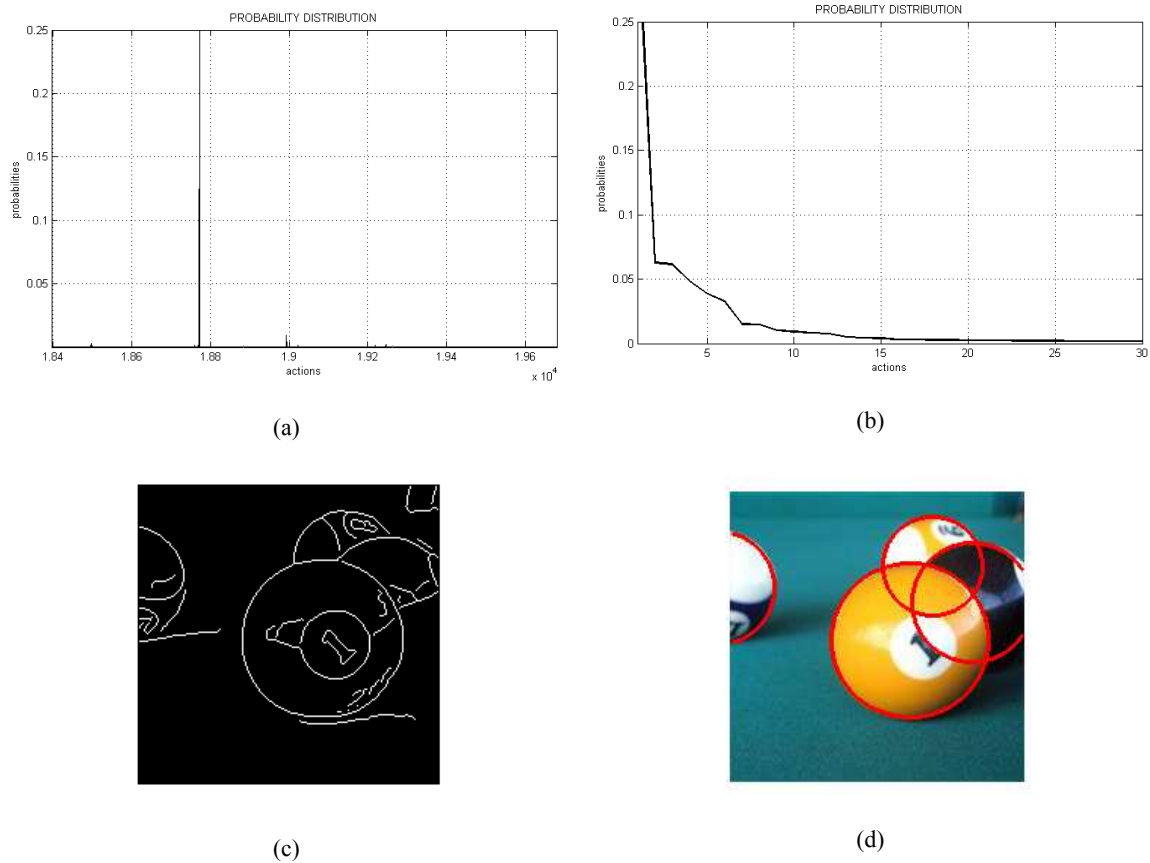

(a)      (b)

(c)      (d)

**Fig. 7.** Overlapped circles. (a) The final probability distribution, (b) the thirty most significant actions, showing the probability in descending order. (c) The edge-map image and (d) its correspondent overlaid detected circles.

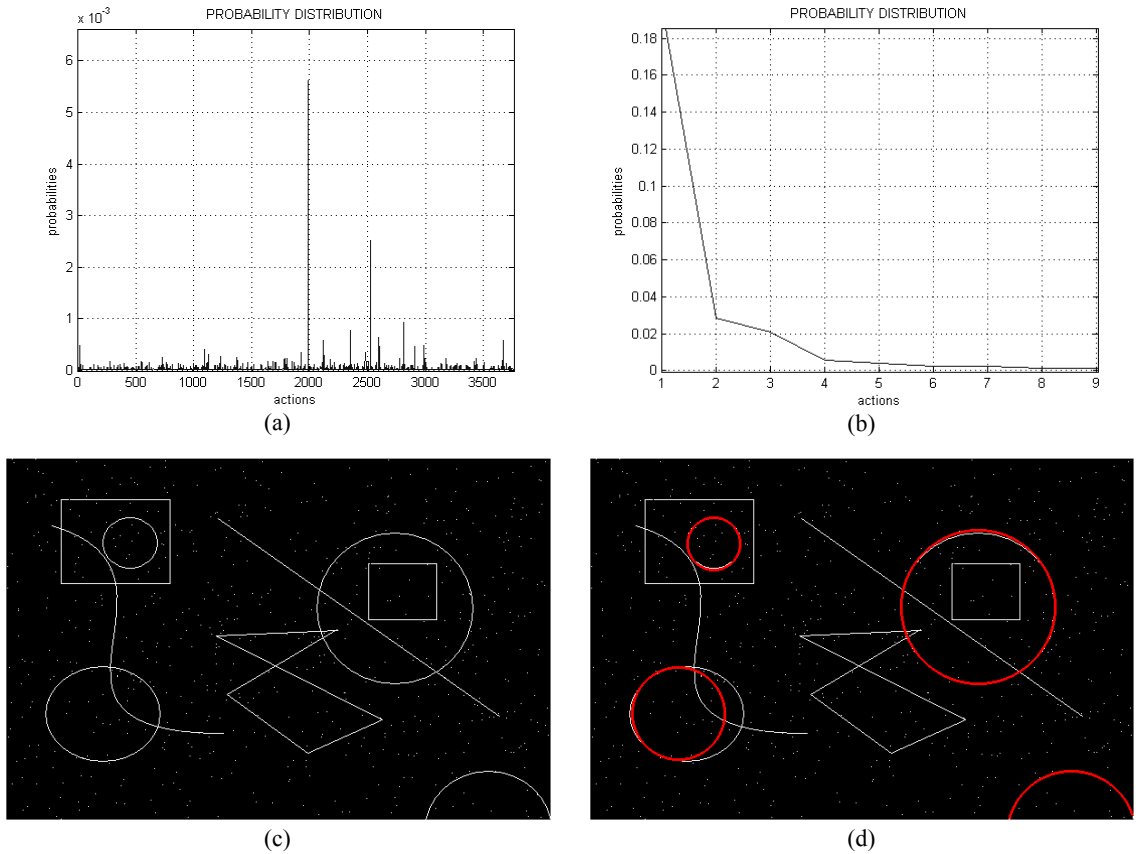

**Fig. 8.** Multiple circle discrimination: (a) the final probability distribution and (b) the nine most significant actions showing the highest probabilities in descending order. (c) The edge-map and (d) its image including detected circles as overlays.

*5.4. Circle discrimination test.*

This section discusses on the algorithm's ability to detect multiple circles despite other shapes are included in the image. Several synthetic images of 400x600 pixels are considered for this experiment. Some of these images are contaminated adding noise to increase the complexity on the detection process. Figure 8 shows the results for one selected image containing nine different shapes with four circular embedded shapes. First two circles (top on the image) are quasi-perfect circles while a third shape (down-left in the image) is an ellipse; finally an occluded circle features on the bottom-right.

*5.5. Performance evaluation.*

In order to extend the algorithm's analysis, the LA algorithm is compared to the BFAOA [16], the GA circle detector [15] and the CSA-based approach [17] over a set of common images.

  **This is a preprint copy that has been accepted for publication in IET Image Processing**

The GA algorithm follows the proposal of Ayala-Ramirez et al. in [15], considering the population size as 70, the crossover probability as 0.55, the mutation probability as 0.10 and the number of elite individuals as 2. The roulette wheel selection and the 1-point crossover are applied. The parameter setup and the fitness function follow the configuration suggested in [15]. The BFAOA algorithm follows the implementation from [16] considering the experimental parameters as: $S=50$, $N_c = 100$, $N_s = 4$, $N_{ed} = 1$, $P_{ed} = 0.25$, $d_{attract} = 0.1$, $w_{attract} = 0.2$, $w_{repellant} = 10$ $h_{repellant} = 0.1$, $\lambda = 400$ and $\psi = 6$. The CSA detector is set using the following parameter values $h=120$, $n=100$, $N_c = 100$, $\rho = 3$, $P_r = 20$, $L=20$, $T_c = 0.01$, $s=8$, $r_{min} = 10$, ITER=400. Such values are found to be the best configuration set according to [17].

Images rarely contain perfectly-shaped circles. In order to test the accuracy for a single-circle, the detection is challenged by a ground-truth circle, which is determined manually from the original edge-map. The parameters $(x_{true}, y_{true}, r_{true})$ of the ground-truth circle are computed considering the best fitted ellipse that a human observer can identify through a drawing software (MSPaint©, etc). If the parameters of the detected circle are defined as $(x_D, y_D)$ and $r_D$, then an error score (Es) is defined as follows:

$$\mathrm{Es} = \eta \cdot (|x_{true} - x_D| + |y_{true} - y_D|) + \mu \cdot |r_{true} - r_D| \tag{11}$$

The central point difference $(|x_{true} - x_D| + |y_{true} - y_D|)$ represents the centre shift for the detected circle as it is compared with the ground-truth circle. The radio mismatch $(|r_{true} - r_D|)$ accounts for the difference between their radii. $\eta$ and $\mu$ represent two weighting parameters, which are to be applied separately to the central point difference and to the radius mismatch for the final error Es. In this study, they are chosen as $\eta = 0.05$ and $\mu = 0.1$. This particular choice ensures that the radii difference would be strongly weighted in comparison to the difference in the central circular positions of manually detected and machine-detected circles. In order to use an error metric for multiple-circle detection, the averaged Es produced from each circle in the image is considered. Such criterion, defined as the multiple error (ME), is calculated as follows:

$$\text{ME} = \left(\frac{1}{NC}\right) \cdot \sum_{R=1}^{NC} \text{Es}_R \qquad (12)$$

where *NC* represents the number of circles actually present the image. In case the ME is less than 1, the algorithm is considered successful; otherwise it is said to have failed in the detection of the circle set. Notice that for $\eta = 0.05$ and $\mu = 0.1$, it yields ME<1, which accounts for a maximal tolerated average difference on radius length of 10 pixels, whereas the maximum average mismatch for the centre location can be up to 20 pixels. In general, the success rate (SR) can thus be defined as the percentage of achieving success after a certain number of trials.

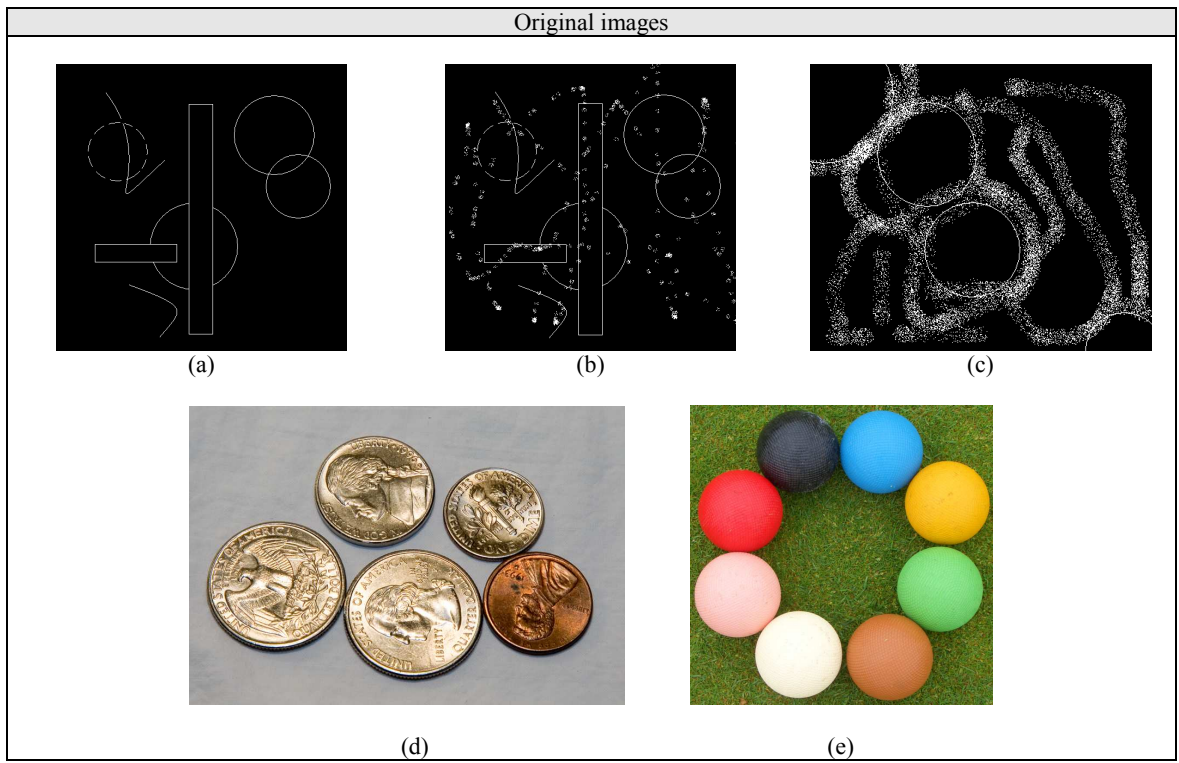

**Fig. 9.** Experimental image set for further testing.

Fig. 9 shows five images that have been used to compare the performance of the GA-based algorithm [15], the BFOA method [16], the CSA-based detector [17] and the proposed approach. Three images ((a), (b) and (c)) of Fig. 9 are synthetic images and some of them have been contaminated by local noise in order to increase the complexity of the detection task. The last two images ((d) and (e)) of Fig. 9 correspond to natural images. Figure 10 shows the outcome after applying each detector over the three synthetic images (a)–(c). Figure 11 presents the experimental results considering the three natural images (d)-(f). For all Figures, performance is analyzed through 35 different executions for each algorithm.

 

Figures report the results for the median-run solutions, i.e. the runs are ranked according to their final fitness values.

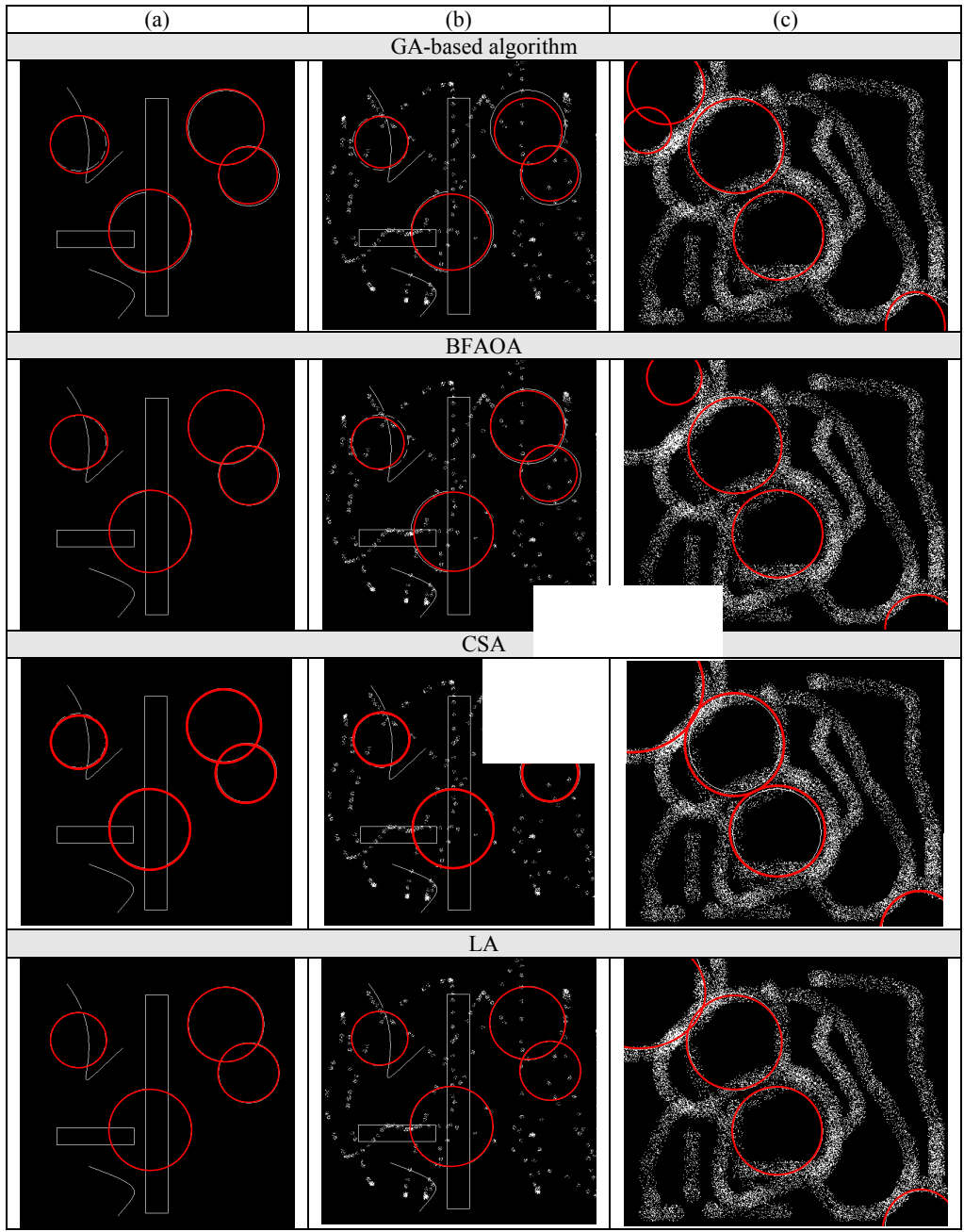

**Fig. 10.** Synthetic images and their corresponding detected circles after applying the GA, the BFAOA, the CSA and the LA algorithms.

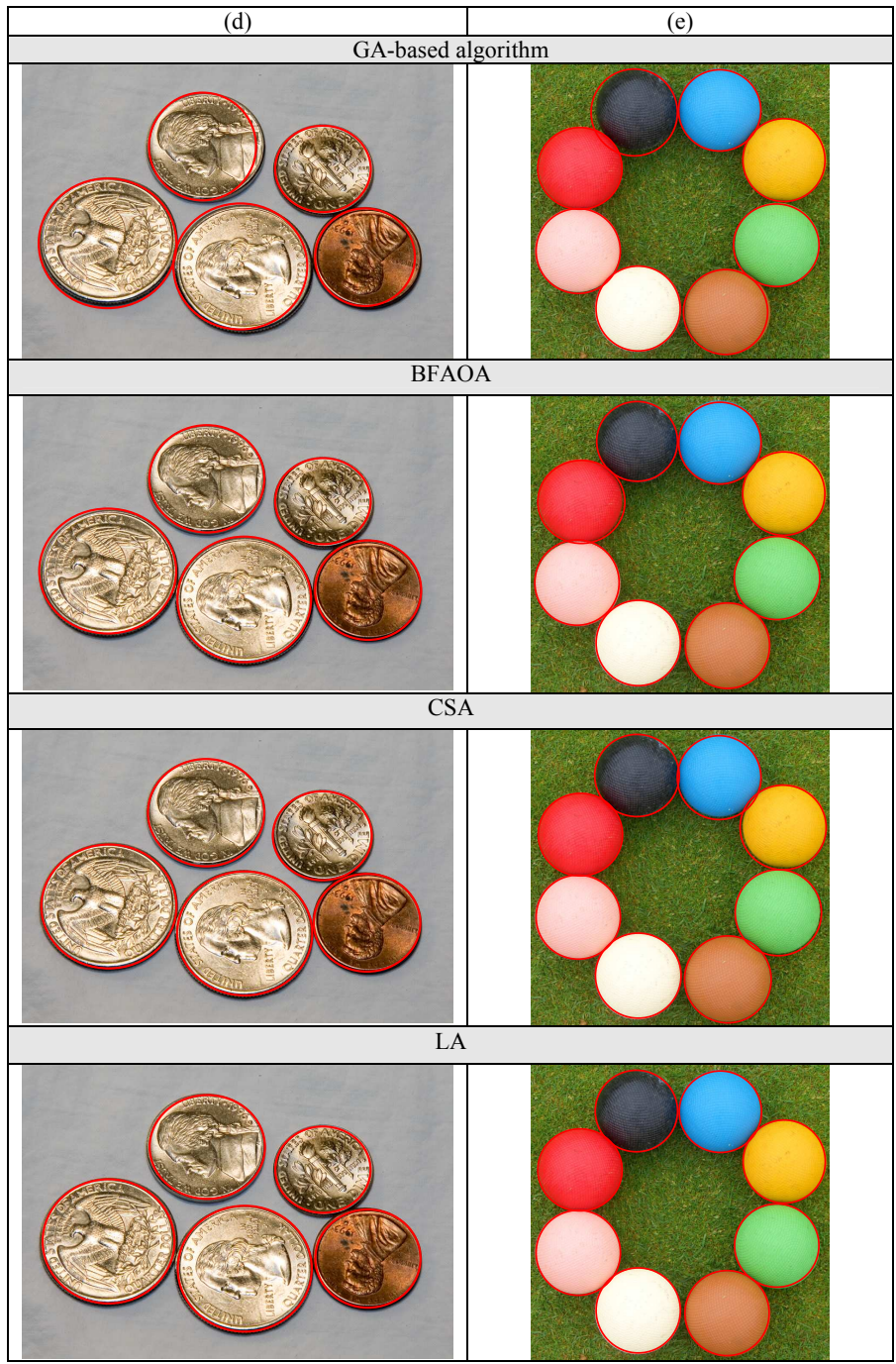

**Fig. 11.** Natural images and their detected circles after applying the GA, the BFAOA, the CSA and the LA algorithms

Table 2 exhibits the success rate (*SR*) in percentage and the averaged multiple error (ME) considering the five test images shown by Fig. 9. Likewise, Table 3 presents the averaged execution time, over the same image set. The performance indexes are determined assuming 35 different executions for each algorithm. The best entries are bold-cased in Table 2 and Table 3. A close inspection over both tables reveals that the proposed method is able to achieve the highest success rate keeping the smallest error yet requiring less computational time for all cases.

| Image | Success rate (*SR*) (%) | | | | Averaged ME ± Standard deviation | | | |
|---|---|---|---|---|---|---|---|---|
| | GA | BFOA | CSA | LA | GA | BFOA | CSA | LA |
| (a) | 90 | **100** | **100** | **100** | 0.76±(0.102) | 0.92±(0.088) | 0.70±(0.075) | **0.41±(0.054)** |
| (b) | 85 | 95 | 98 | **100** | 0.87±(0.095) | 0.95±(0.077) | 0.71±(0.029) | **0.34±(0.029)** |
| (c) | 87 | 91 | 95 | **99** | 0.89±(0.115) | 0.97±(0.023) | 0.68±(0.077) | **0.28±(0.032)** |
| (d) | 80 | 92 | 98 | **100** | 0.88±(0.074) | 0.98±(0.018) | 0.57±(0.086) | **0.38±(0.023)** |
| (e) | 71 | 88 | 98 | **100** | 0.92±(0.035) | 1.18±(0.076) | 0.63±(0.051) | **0.30±(0.039)** |

**Table 2.** The success rate (*SR*) and the averaged multiple error (ME) for the GA-based algorithm, the BFOA method, the CSA approach and the proposed LA, considering the five test images shown by Fig. 9.

| Image | Averaged execution time (*s*) ± Standard deviation | | | |
|---|---|---|---|---|
| | GA | BFOA | CSA | LSA |
| (a) | 4.55±(0.32) | 3.90±(0.23) | 1.21±(0.41) | **0.12±(0.15)** |
| (b) | 3.66±(0.37) | 2.98±(0.37) | 0.98±(0.27) | **0.16±(0.13)** |
| (c) | 3.82±(0.42) | 4.22±(0.21) | 1.17±(0.31) | **0.15±(0.11)** |
| (d) | 8.1±(0.51) | 9.21±(0.22) | 2.98±(0.72) | **0.21±(0.21)** |
| (e) | 6.88±(0.60) | 12.78±(0.43) | 3.21±(0.53) | **0.19±(0.29)** |

**Table 3.** The averaged execution time for the GA-based algorithm, the BFOA method, the CSA approach and the proposed LA algorithm, considering the five test images shown by Fig. 9.

In order to statistically analyze the results in Table 2, a non-parametric significance proof known as the Wilcoxon's rank test [31,32] has been conducted. Such proof allows assessing result differences among two related methods. The analysis is performed considering a 5% significance level over multiple error (ME) data. Table 4 reports the *p*-values produced by Wilcoxon's test for a pair-wise comparison of the multiple error (ME), considering three groups gathered as LA vs. GA, LA vs. BFOA and LA vs. CSA. As a null hypothesis, it is assumed that there is no difference between the values of the two algorithms. The alternative hypothesis considers an existent difference between the values of both approaches. All *p*-values reported in the Table 4 are less than 0.05 (5% significance level) which is a strong evidence against the null hypothesis, indicating that the LA averaged values for the performance are statistically significant which has not occurred by chance.

| Image | *p*-Value | | |
|---|---|---|---|
| | LA vs. GA | LA vs. BFOA | LA vs. CSA |
| (a) | 1.3421e-004 | 1.0124e-004 | 2.3527e-004 |
| (b) | 1.4211e-004 | 1.0328e-004 | 2.2781e-004 |
| (c) | 1.1267e-004 | 1.1029e-004 | 2.3486e-004 |
| (d) | 1.2327e-004 | 1.0345e-004 | 2.7397e-004 |
| (e) | 1.3260e-004 | 1.1003e-004 | 2.8826e-004 |

**Table 4.** *p*-values produced by Wilcoxon's test comparing LA to GA, BFOA and CSA over an averaged multiple error ME (Table 3).

*5.6. Incremental noise.*

In this section, the ability of the LA algorithm to detect circles in nosy images is compared. The test considers images of 500x400 pixels containing three well characterized circles with prior defined centers and radii. The images are contaminated with ten different levels of salt & pepper noise from 0.01 to 0.10 (see Figure 12). For comparison purposes, the LA algorithm is tested against the RHT [12] and the IRHT [14] circle detectors. For the RHT algorithm, the parameter values are defined as suggested in [12]. However, in the case of IRHT, the most important parameters are grouped into the vector $\Delta_c$ which defines the desired set of enlargements of the circle/ellipse parameters to build a new region of interest. In this comparison, $\Delta_c$ has been chosen as $[0.5 \cdot \sigma_x \quad 0.5 \cdot \sigma_x \quad 0.5 \cdot \sigma_a \quad 0.5 \cdot \sigma_b \quad 0]$, assuming that according to [34] those values yield a reduced sensitivity to noisy images. Table 5 presents the multiple error (ME, see Eq. 11 and 12) values averaged over one hundred executions of each algorithm. It can be seen that although the RHT and IRHT algorithms demonstrate an acceptable performance over images featuring low noise levels, the ME values increase significantly as the noise increases.

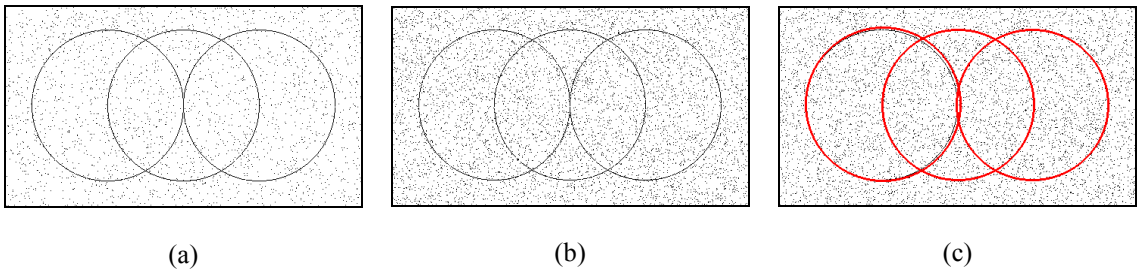

(a)          (b)          (c)

**Fig. 12.** Incremental noise: (a) and (b) show images with different noise level that have been used as inputs for the experiment and (c) shows one of such images including some detected circles as overlays.

| Noise Level | Averaged ME values | | |
|---|---|---|---|
| | LA | RHT | IRHT |
| 0.01 | 0 | 0 | 0 |
| 0.02 | 0 | 0 | 0 |
| 0.03 | 0 | 0 | 0 |
| 0.04 | 0 | 0 | 0 |
| 0.05 | 0.30 | 0 | 0 |
| 0.06 | 0.32 | 0 | 0 |
| 0.07 | 0.39 | 12.34 | 11.21 |
| 0.08 | 0.37 | 14.32 | 20.12 |
| 0.09 | 0.51 | 17.67 | 22.87 |
| 0.10 | 0.68 | 20.64 | 24.34 |

**Table 5.** Comparison between the LA-based, the RHT and the IRHT methods while processing images at Figure 12

## 6. Conclusions

This paper has presented an algorithm for the automatic detection of multiple circular shapes from complicated and noisy images with no consideration of the conventional Hough transform principles. The detection process is considered to be similar to an optimization problem. As it differs from other heuristic methods that employ an iterative method, the proposed method is able to detect single or multiple circles over a digital image considering only one optimization procedure. The LA algorithm searches the entire edge-map for circular shapes by using a combination of three non-collinear edge points as candidate circles (actions) in the edge-only image. A reinforcement signal (matching function) is used to measure the existence of a candidate circle over the edge map. Guided by the values of this reinforcement signal, the set of encoded candidate circles are evolved using the LA so that the best candidate can fit into an actual circle. The probability distribution is analyzed after the optimization process is executed in order to find other local minima, i.e. other circles. The approach generates a fast sub-pixel detector which can effectively identify multiple circles in real images despite circular objects exhibiting a significant occluded portion.

Classical Hough Transform methods for circle detection use three edge points to cast a vote for the potential circular shape in the parameter space. However, they would require a huge amount of memory and longer computational times to obtain a sub-pixel resolution. Moreover, HT-based methods rarely find a precise parameter set for a circle in the image [33]. In our approach, the detected circles hold a sub-pixel accuracy inherited directly from the circle equation and the MCA method.

In order to test the circle detection performance, both speed and accuracy have been compared. Score functions are defined by (11)-(12) in order to measure accuracy and effectively evaluate the mismatch between manually-detected and machine-detected circles. We have demonstrated that the LA method outperforms the GA approach (as described in [15]), the BFOA method (as described in [16]) and the CSA detector (as described in [17]) within a statistically significant framework (Wilcoxon test). In contrast to the LA method, the RHT algorithm [12] and the IRHT [14] detector show a decrease in performance under noisy conditions. Yet the LA algorithm holds its performance under the same circumstances.

Although Table 2 and Table 3 indicate that the LA method can yield better results on complicated and noisy images in comparison to the GA, the BFAOA and the CSA methods, the aim of our paper is not intended to beat all the circle detector methods proposed earlier, but to show that the LA algorithm can effectively serve as an attractive method to successfully extract circular shapes.